\documentclass[conference]{IEEEtran}
\IEEEoverridecommandlockouts
\usepackage{cite}
\usepackage{amsmath,amssymb,amsfonts}
\usepackage{algorithmic}
\usepackage{graphicx}
\usepackage{color}
\usepackage{textcomp}
\usepackage{fourier} 
\usepackage{array}
\usepackage{makecell}
\usepackage{flushend}
\usepackage{tikz}

\def\BibTeX{{\rm B\kern-.05em{\sc i\kern-.025em b}\kern-.08em
    T\kern-.1667em\lower.7ex\hbox{E}\kern-.125emX}}

\def\mycopyrightnotice{%
  \begin{minipage}{\textwidth}
  \small
  Copyright~\copyright~2019 IEEE. Personal use of this material is permitted. Permission from IEEE must be obtained for all other uses, in any current or future media, including reprinting/republishing this material for advertising or promotional purposes, creating new collective works, for resale or redistribution to servers or lists, or reuse of any copyrighted component of this work in other works by sending a request to pubs-permissions@ieee.org.
  
  \vspace{1.0cm}
  Accepted and published in: Proceedings of the 2019 International Conference on Robotics and Automation (ICRA), 2019, pp. 7441-7447, doi: 10.1109/ICRA.2019.8793801.
  
  \vspace{1.0cm}
  This is the preprint only. The final version of the paper is available at: https://ieeexplore.ieee.org/document/8793801
  
  \vspace{1.0cm}
  Cite this paper as:
  
  \vspace{0.5cm}
  D. Belter, J. Bednarek, H. -C. Lin, G. Xin and M. Mistry, "Single-shot Foothold Selection and Constraint Evaluation for Quadruped Locomotion," 2019 International Conference on Robotics and Automation (ICRA), 2019, pp. 7441-7447, doi: 10.1109/ICRA.2019.8793801
  \end{minipage}
}
\makeatother
    
\begin{document}
\mycopyrightnotice

\title{Single-shot Foothold Selection and Constraint Evaluation for Quadruped Locomotion\\
\thanks{This research was supported by EU Horizon 2020 project THING.}
}

\author{\IEEEauthorblockN{Dominik Belter, Jakub Bednarek & Hsiu-Chin Lin$^*$, Guiyang Xin$^*$, Michael Mistry}
\IEEEauthorblockA{\textit{Institute of Control, Robotics and Information Engineering} & \textit{School of Informatics}\\
\textit{Poznan University of Technology} & \textit{University of Edinburgh}\\ 
Poznan, Poland & Edinburgh, UK\\}
\thanks{*The authors contributed equally}
\thanks{contact: dominik.belter@put.poznan.pl, jakub.bednarek@put.poznan.pl}
}

\maketitle

\begin{abstract}
In this paper, we propose a method for selecting the optimal footholds for legged systems. The goal of the proposed method is to find the best foothold for the swing leg on a local elevation map. We apply the Convolutional Neural Network to learn the relationship between the local elevation map and the quality of potential footholds. 
The proposed network evaluates the geometrical characteristics of each cell on the elevation map, checks kinematic constraints and collisions.  
During execution time, the controller obtains the qualitative measurement of each potential foothold from the neural model.
This method allows to evaluate hundreds of potential footholds and check multiple constraints in a single step which takes 10~ms on a standard computer without GPGPU. The experiments were carried out on a quadruped robot walking over rough terrain in both simulation and real robotic platforms.
\end{abstract}


\section{Introduction}
\noindent 
Locomotion in challenging terrain requires careful selection of footholds. The robot should select a stable support for each foot to avoid slippages and falls. This strategy is crucial when the robot needs to deal with highly irregular terrain. A challenging example is the ``crossing the stream'' problem where the robot has to reach another bank of the river over a sequence of stones which are the only acceptable footholds. A poor foothold selection method means that the robot may fall into the river. 

In contrast, other types of locomotion assume that the robot walks dynamically on rough terrain and stabilizes its posture using fast control algorithms and compliant legs~\cite{Hutter2016,Hyun2014}. This approach is efficient on moderately rough terrain, which means that the robot can quickly reach the goal position. In this case, the robot does not have to build the precise model of the terrain. The problem of stable locomotion is solved by the controller which reacts to disturbances like slippages resulting from unstable footholds. To efficiently navigate in the unknown environment, perception systems are required to detect high obstacles so the robot can avoid them while walking~\cite{Wooden2010}. However, this type of locomotion will fail when the robot has to face the ``crossing the stream'' problem.

\begin{figure}[t]
 \centering
\includegraphics[width=0.8\columnwidth]{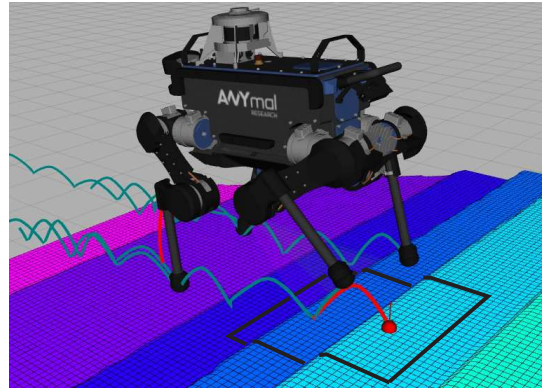}
\caption{The foothold selection problem for a quadruped robot. The region of the elevation map below the $i$-th leg is evaluated to find the best foothold. Each candidate position of the foot has to be kinematically feasible and collision-free. All coordinate systems are defined in the world frame $W$.}
 \label{footholdMap}
\end{figure}

The problem of foothold selection is presented in Fig.~\ref{footholdMap}. 
The robot evaluates the region (elevation map) below the $i$-th leg to select the best foothold that fulfills the following requirements. First, the robot should avoid selecting footholds on sharp edges and/or steep slopes because they are potentially risky. 
Second, the selected foothold should be within the kinematic limit of the robot (inside the workspace of the leg). 
Also, the robot should avoid self-collision and check whether the thigh or shank collides with the terrain.
In the classical approach, all constraints are verified sequentially by the controller of the robot~\cite{Belter2011}. 

The deliberative approach for locomotion over rough terrain requires a good model of the terrain. 
A full 3D model of the environment can be obtained using terrain mapping methods, such as OctoMap~\cite{Hornung2013} or Normal Distribution Transform Occupancy Maps (NDT-OM)~\cite{Saarinen2013}. 
However, the computation and memory requirements increase as the resolution of the 3D map increases, and the controller has to analyze hundreds of potential footholds at each time-step, which make it difficult in real-time robot control.
In contrast, the elevation map sufficiently represents the terrain and guarantees quick access to each cell.
Also, the elevation map can be directly transformed to grayscale image to feed the Convolutional Neural Network~\cite{Krizhevsky2012} which we use in this research to select the best foothold.

In this research, we propose a computationally efficient solution for foothold selection. 
We applied a neural network to learn a model (off-line) that maps the properties of the terrain to the quality of a potential foothold while excluding footholds which are risky or kinematically infeasible.
During execution time, we efficiently predict the quality of a potential foothold from the learned model. 
The proposed method is verified on a quadruped robot walking over a rough terrain, in both simulation and real robot platforms.

\section{Related Work}
\noindent
The problem of foothold selection is similar to the problem of multi-finger grasping and was studied widely by the robotics community. Recent development in this field includes the method which use local geometrical properties of the objects to find the acceptable positions of the fingertips on the object's surface~\cite{Kopicki2015}. The grasp configurations are trained from the real examples. The collision and kinematic constraints are taken into account during the inference procedure. 
Recently, deep neural network, such as the Convolutional Neural Networks (CNN) gained high popularity in robotics applications. In grasp, the CNN is applied to select feasible grasp and robotics finger positions on the object's surface using point cloud~\cite{Mahler2017} or depth images\cite{Gualtieri2016}.

Before the deep neural networks were applied, most approaches for the foothold selection were based on the local features computed for the terrain surface. For example, the Ambler robot computes the inclination of the terrain, roughness, and local curvature from the elevation maps~\cite{Krotkov1996}. These features are provided to the input of the simple neural network which was trained on the data provided by human experts. 
Another method which uses the elevation map was proposed by Chen and Kumar~\cite{Chen2009}. The method estimates a probability map that is related to the capability of each cell to provide stable support for the robot's feet.
Another unique method for foothold assessment is proposed by~\cite{Hoepflinger2013}. Their robot is equipped with the haptic device on the feet, which explores and evaluates the potential footholds without human supervision. 
The controller of the HyQ robot is focused more on the reflexes which stabilize the robot~\cite{Focchi2018}. The visual information about the terrain is used to place the foot on the terrain surface without avoiding risky footholds~\cite{Barasuol2015}. 
The robot corrects the nominal foothold positions according to the output from the visual pattern classifier applied on the terrain patches. The map is built on-line using the RGB-D sensor.

A great progress in the field of autonomous legged locomotion on rough terrain was done on the quadruped robot LittleDog. 
Rebula {\em et al.} \cite{Rebula2007} proposed a terrain scorer which computes the spatial relationship between a considered point and its neighboring points and then rejects points which are located on edges, large slope, the base of a cliff, or inside of a hole. 
A learning-based method was proposed to evaluate terrain templates based on the human demonstration~\cite{Kalakrishnan2009}. 
The terrain scorer approach is also adapted in~\cite{Kolter2008}, where the weights of geometric features of the terrain are obtained during training and then used for the footsteps planning.

The foothold selection method for a six-legged robot is represented by the method implemented on the Lauron IV robot~\cite{Roennau2009}. The foothold selection module considers points around initial foothold and takes into account elevation credibility, the mean height, and the height variance of the cells. 
Previous work on a six-legged robot, Messor, learns which points on the elevation map can provide stable support from simulation data~\cite{Belter2011}. 
Then, the trained Gaussian Mixture is used to select the footholds in the RRT-based motion planner~\cite{Belter2016jfr}. The kinematic and self-collision constraints are also taken into account; however, this process significantly slow-downs the foothold selection process.

\subsection{Approach and Contribution}
In this paper, we propose a novel method to evaluate potential footholds for the quadruped robot in a single step using Convolutional Neural Network. 
We collect data for training the network using the kinematic model of the robot and elevation map of a rough terrain. 
The network implicitly stores information about the kinematic model of the leg and can detect footholds which are outside the workspace.
The network takes into account kinematic and collision constraints.
The trained model also rejects footholds which are not collision-free. We are the first who show that the Convolutional Neural Network can be used to evaluate geometrical properties of the potential positions (footholds) on the map and simultaneously consider all kinematic constraints which are related to the model of the robot. Therefore, the resulting footholds are feasible. 

In contrast to~\cite{Krotkov1996}, the method based on CNN does not depend on the manually computed features but extracts features automatically from data. 
Our method evaluates 400 potential footholds and checks constraints in a single inference step. In contrast to previous work, our new approach only needs approximately 10~ms on the CPU.

\section{Foothold Selection Module}

\begin{figure}[t]
 \centering
\includegraphics[width=0.89\columnwidth]{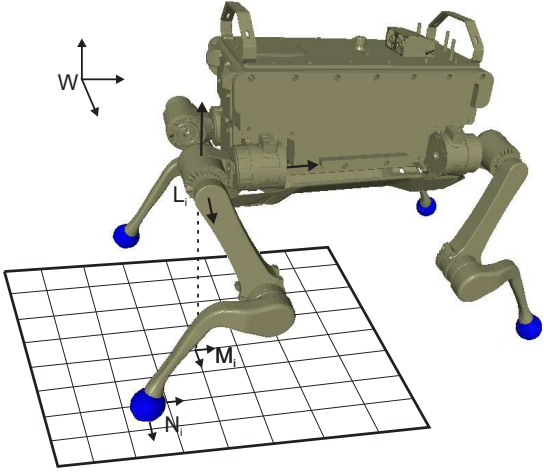}
\caption{Center of the local map $M_i$ used for foothold selection is located below the $i$-th leg joint $L_i$. The foothold selection algorithm considers also the nominal position of the foot $N_i$.}
 \label{localMap}
\end{figure}

\noindent We propose a foothold selection module to evaluate potential footholds which are inside the local map extracted from the elevation map build by the robot~\cite{Fankhauser2014}. 
The size of the global map build by the robot is $6\times6 m^2$ and the size of each cell is $2\times2~cm^2$. The center of the map is defined by the center of the robot projected on the ground. 

The {\em local map} is presented in Fig.~\ref{localMap}. The center of the local map $M_i$ is defined by a point below the joints of the considered leg. The size of the local map is 40$\times$40 cells, and this is set to cover the kinematic range of the leg. 
We also consider the {\em nominal foothold} $N_i$, the desired foothold assuming that the robot is walking on a flat terrain, as part of the criteria for selecting the optimal foothold.

In the following sections, we describe our approach on data gathering, offline training, and online inference for foothold selection.

\subsection{Dataset}

\begin{figure}[t]
 \centering
\includegraphics[width=0.99\columnwidth]{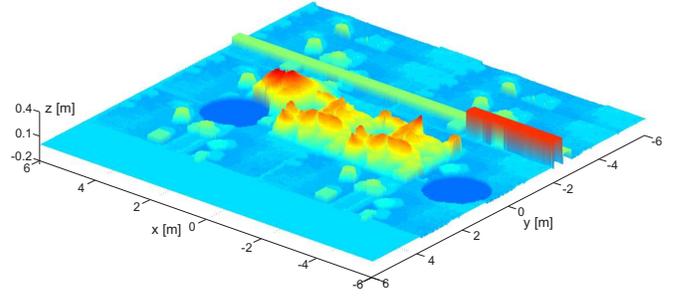}
\caption{Elevation map used to collect data for training.}
 \label{trainMap}
\end{figure}

To train the neural network we collect the samples on the elevation map presented in Fig.~\ref{trainMap}. We collect the data for two legs only and train two separate models. The ANYmal robot used in this research is symmetrical and we can use the same model to evaluate foothold for the right and left legs. To this end, we have to flip horizontally the input terrain map and after inference, we flip horizontally the obtained cost map.

To generate data for training, we randomly select the position of the robot on the map (horizontal position and distance to the ground). The orientation of the robot on the horizontal plane (yaw angle) is randomly selected from four main orientations: $n\cdot \frac{\varPi}{2}$, for $n=0,1,2,3$. For the obtained pose of the robot, we compute the pose of the $i$-th leg and we extract local 40$\times$40 map. 

For the input map, we compute the desired output map. For each cell of the map we check five constraints:
\begin{itemize}
 \item kinematic range,
 \item self-collision,
 \item collisions with the ground,
 \item kinematic margin,
 \item cost related to the local shape of the terrain.
\end{itemize}

\noindent First, we check the kinematic range. If the given position of the foot is outside the workspace of the considered leg we set the cost to the maximal value (255). Then we check self-collisions and collisions with the ground. To determine if the parts of the robot collide between each other we create a mesh model of each part and use Flexible Collision Library to determine collsions~\cite{Pan2012}. The same procedure is performed to check collision with the ground. Of course, we do not consider collisions between the foot and the terrain model. If we detect collision the cost of the foothold is set to 255 and we do not check other constraints.

\begin{figure*}[t]
 \centering
\includegraphics[width=1.99\columnwidth]{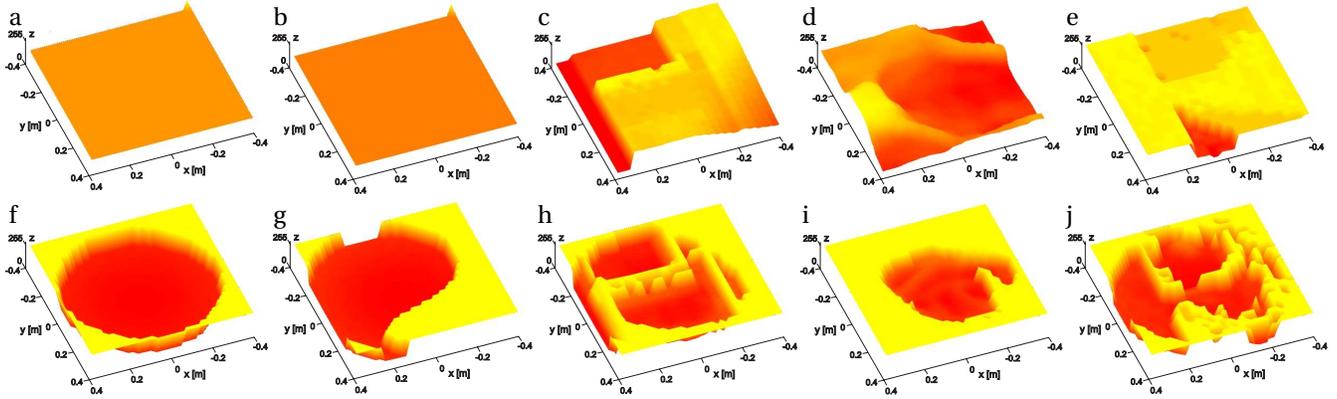}
\put(-500,140){a} \put(-400,140){b} \put(-300,140){c} \put(-200,140){d} \put(-100,140){e} 
\put(-500,65){f} \put(-400,65){g} \put(-300,65){h} \put(-200,65){i} \put(-100,65){j}
\caption{Example training data: local elevation maps (a,b,c,d,e), and corresponding terrain cost (f,g,h,i,j)}
 \label{train}
\end{figure*}
\noindent 

If the given position of the foothold is collision-free and the foot is inside the workspace, we compute the kinematic margin $c_k$. The kinematic margin is the distance between the current position of the foot and the border of the workspace. If the kinematic margin is equal to 0, then the leg cannot move in one direction. The maximal value of the kinematic margin means that the leg has the maximal motion range. Finally, we compute the cost which depends on the properties of the local map. We use the function which computes the cost $c_m$ for the hexapod Messor robot~\cite{Belter2011}. We can use the relation obtained for the Messor robot because both robots have the same hemispherical feet. The function obtained~\cite{Belter2011} computes the local terrain cost and does not depend on the kinematic model of the robot. Finally, we compute the final cost of the considered foothold $c_f$ and scale the cost to the range $[0,255]$:

\begin{equation}
 \label{costFoot}
c_{f} = \frac{c_k+2 \cdot c_m}{3} \cdot 255.
\end{equation}

\begin{figure}[t]
 \centering
\includegraphics[width=0.99\columnwidth]{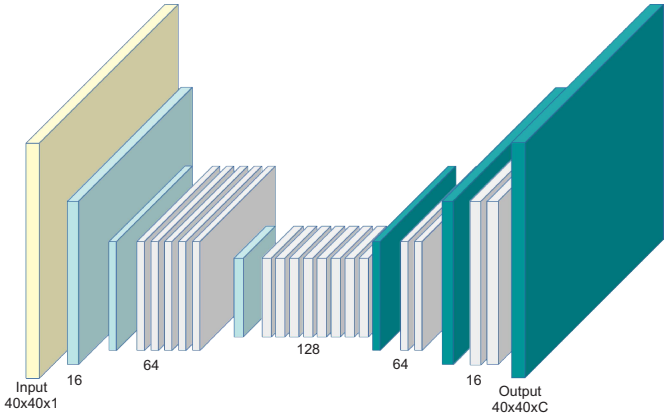}
\caption{The model of ERF network. Light blue blocks represent downsampling, dark blue - upsampling by transposed convolution and white blocks show residual layers. Numbers below blocks describes the number of feature maps used in specific levels. $C$ denotes the number of classes (14 in the current implementation).}
 \label{ERF}
\end{figure}

\noindent We repeat the procedure for each cell of the input elevation map and save the input (elevation map) and the output (terrain) cost images. We collected 20000 training pairs for each leg. The example training data are presented in Fig.~\ref{train}. First two columns present the computed cost maps for the flat terrain. In this case, the output depends on the leg's workspace and the kinematic margin. The distance between the terrain and the robot is larger on the map in Fig.~\ref{train}a than in Fig.~\ref{train}b. The obtained cost maps (Fig.~\ref{train}f and Fig.~\ref{train}g) represents the horizontal cross-section over the workspace of the robot's leg. The yellow cells represent positions of the foot which are outside the workspace and are inaccessible for the robot. The red cells are correlated to the value of the kinematic margin. In the following examples in Fig.~\ref{train}c--e the terrain is irregular and we can observe how the workspace of the robot is limited by the terrain shape. The edges on the obstacles significantly increase the cost of footholds.

\begin{figure*}[t]
 \centering
\includegraphics[width=1.99\columnwidth]{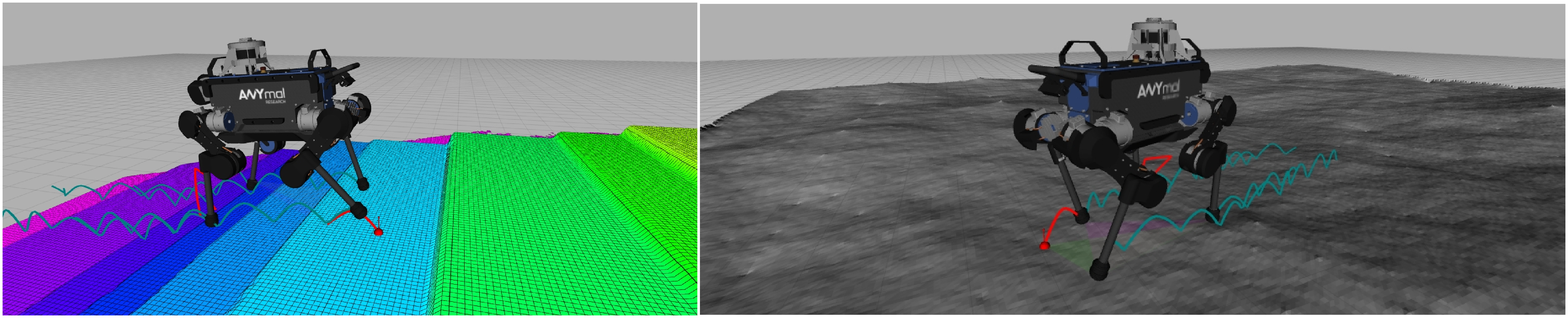}
\put(-500,92){a} 
\put(-275,91){b}
\caption{Experiment with the ANYmal robot on stairs (a) and on rough terrain (b) in the Gazebo simulator}
 \label{expRobot}
\end{figure*}

\subsection{Convolutional Neural Network}
\noindent 
The neural network should extract features from local elevation map and evaluate the potential footholds. We select an architecture that can be run in real-time on the machine without GPGPUs. Because the neural networks are much more efficient in solving the classification than the regression problem we discretize input cost and divide training examples into $C$ different classes. The neural network returns the id of the class which is related to the predicted terrain cost.

The proposed CNN architecture is an Efficient Residual Factorized ConvNet (ERF) first introduced in~\cite{ERF}. The characteristics of this model is the modification of the \textit{residual layer}~\cite{ResNet} called \textit{residual non-bottleneck 1D layer}. The 2D convolution with filter shape 3$\times$3 is replaced by two 2D convolutions with filter shapes 3$\times$1 and 1$\times$3. This approach reduces the number of variables and complexity. The ERF model is shown in Fig.~\ref{ERF}. First, the input data is processed twice by downsampling blocks. The downsampling blocks are created from the concatenation of the max pooling and 2D convolution with filter shape 3$\times$3, and the stride set to 2. The concatenation is followed by the activation function. Then, five residual layers and another downsample block are added. The output of the encoder part is processed by eight residual layers which are interwoven with different dilation rate applied to the convolutions. The decoder part of the model consists of two series of convolutional upsample and two residual layers. The upsampling is performed by transposing convolution with the stride set to 2. The output of the model is produced by upsampling convolution with filter shape 2$\times$2 and number of filters equal to the number of classes and stride 2. Activation function used in each nonlinear layer is a rectified linear unit (\textit{ReLU}).

The optimized objective of the model is composed of cross-entropy loss and regularization loss. The cross-entropy is additionally weighted according to the ground truth number of examples for each class label in the training dataset. We computed statistics for the training datasets to find weights related to each class like in~\cite{ENet}. 

\begin{equation}\label{weighting}
w_{i} = \frac{1}{\log(c + p_{i})}
\end{equation}

\noindent where $w_{i}$ is the weight of the i-th class, $c$ is a constant and $p_{i}$ is a probability of the occurrence of the $i$-th class based on the share of examples marked as the $i$-th in the entire training dataset. In this application we used $c=1.08$. The weighting of the cross-entropy allows the handling of unbalanced data. In the training dataset, the most examples are provided for the class which represents footholds inaccessible for the robot (class id 255). Without balancing the training data the network learns to recognize well the predominant class and very often misclassifies the remaining classes.

We use the method presented in~\cite{ADAM} for training the model with an initial learning rate of 5e-4. Additionally, the exponential decay was applied after each epoch to the learning rate with a factor of 0.98. Because of the nature of the training examples, we can't use any of the known data augmentation methods. In order to measure the quality of models accuracy and intersection over union (IoU) metrics were calculated. The learning process took place in 500 epochs. 
The results obtained by two ERF models for front and rear legs are shown in Tab.~\ref{tab:erf_metrics}.

\begin{table}[ht]
\centering
\begin{tabular}{c|c|ccc}
Leg & Accuracy [\%] & IoU \\ \hline
front leg & 82.61 & 49.9 \\ 
rear leg & 82.61 & 49.88\\ 
\end{tabular}
\caption{Accuracy and Intersection over Union (IoU) obtained on validation set for front and rear leg models}
\label{tab:erf_metrics}
\end{table}

\begin{figure}[t]
 \centering
\includegraphics[width=0.89\columnwidth]{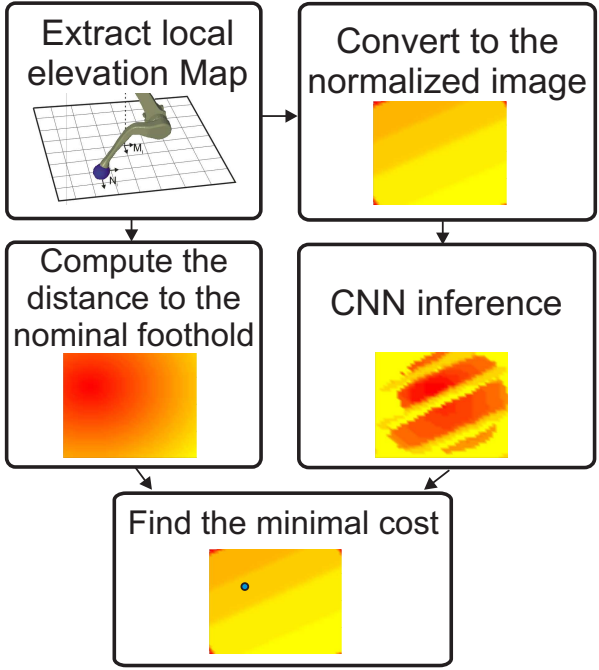}
\caption{Foothold selection procedure on the local elevation map with the Convolutional Neural Network}
 \label{inferenceBlock}
\end{figure}

\subsection{Inference procedure}

\noindent The inference procedure is presented in Fig.~\ref{inferenceBlock}. In the first step, we get submap from the global map built by the robot. The obtained map is aligned with the world coordinate system $W$ but our neural network uses the elevation map which is aligned with the robot coordinate system. Thus, we rotate the obtained local map by the current orientation of the robot on the horizontal plane (yaw angle).
Some information about cells at the corners loses during this rotation, therefore, we take a slightly larger map for rotation purpose. 
Before rotation, the size of the local map is 51$\times$51 cells and after rotation, we crop the map to size 40$\times$40 cells.

In the next step, we convert the obtained elevation map to the image. To this end, we compute the distances between the $i$-th leg coordinate system $L_i$ and each cell of the map. We use 8-bit grayscale images as an input to the network so the obtained distance values are fitted into range 0--255. We use constant normalization factor (0.85~m) for each leg of the robot. The distance which is smaller than the normalization factor is represented by the minimum value in the image (0), and the distance which is larger than the normalization factor is represented by the maximal value in the image (255). The obtained image which represents the terrain patch around the consider leg is the input to the neural network model. 

The network classifies each pixel on the image, and the cost at each pixel corresponds to the cost of taking that foothold. The example inference results for the input image representing stairs are presented in Fig.~\ref{inferenceBlock}. The pixels which are located on the edges between steps on the output image are brighter which means that the robot should avoid these footholds. At the same stage of the inference procedure, we compute the distance from the nominal foothold $d_n$. Then, we compute the final cost $c_{\rm final}$ for each pixel (foothold):

\begin{equation}
 \label{costFinal}
c_{\rm final} = Z_c + k \cdot d_n,
\end{equation}

\noindent where $k$ is the constant value which determines the influence of the distance from the nominal foothold on the final cost of the potential foothold. In the experiments presented in the paper the $k$ value is set to 160. We compute the final cost $c_{\rm final}$ for each pixel on the image (c.f. Fig.~\ref{inferenceBlock}) and we find the minimal value. Then, the pixel with the minimal cost in image coordinates are converted into 3D points in the world coordinate. The obtained value is sent to the controller which executes the motion for the given foothold.

\section{Results}

\noindent First experiments are performed in the Gazebo simulator. We verified the proposed foothold selection method on the ANYmal robot walking on stairs (Fig.~\ref{expRobot}a) and on rough terrain (Fig.~\ref{expRobot}b). The robot uses simulated Intel RealSense D435 RGB-D sensor to build a map of the environment~\cite{Fankhauser2014}. We use the controller presented in~\cite{Fankhauser2018} to plan the foot trajectories above the obstacles, estimate the state of the robot and execute planned trajectories. We only replaced the foothold selection model.

\begin{figure}[t]
 \centering
\includegraphics[width=0.99\columnwidth]{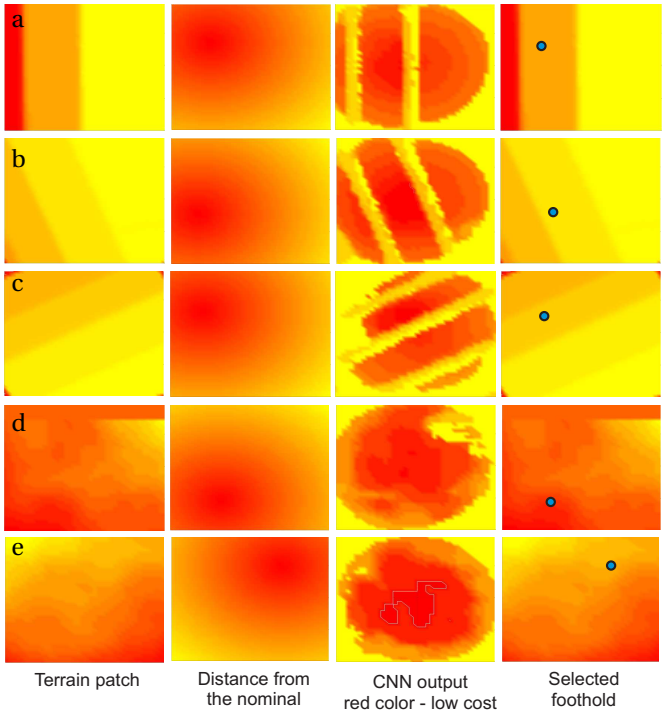}
\put(-246,265){a} 
\put(-246,212){b}
\put(-246,165){c}
\put(-246,112){d}
\put(-246,65){e}
\caption{Example inference results obtained during the experiment on stairs (a,b,c) and on rough terrain (d,e)}
 \label{resultsGazeboExp1}
\end{figure}

The example inference results are presented in Fig.~\ref{resultsGazeboExp1}. We provide the terrain patches extracted from the global elevation map, the distance between potential footholds and the nominal foothold, the output from the CNN, the selected foothold. It is clearly visible from the result obtained on the stairs that the robot avoids placing its feet on the edges. These regions are classified by the neural network as risky and rejected by the foothold selection module. The similar behavior can be observed in the results obtained on rough terrain. In this case, the obstacles are more irregular. For both patches obtained on rough terrain, the region in the center of the workspace has the similar cost. In this case, the distance from the nominal foothold plays an important role. The selected foothold is close to the nominal foothold but still on the position with acceptable foothold cost predicted by the CNN.

\begin{figure}[t]
 \centering
\includegraphics[width=0.8\columnwidth]{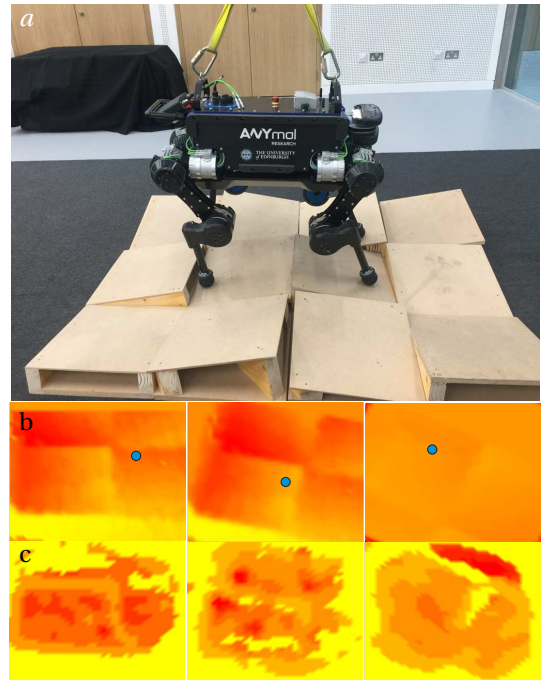}
\put(-198,248){\color{white}{\it a}} 
\put(-198,95){b}
\put(-198,45){c}
\caption{Experiment with the ANYmal robot on the rough terrain mockup (a): example terrain patches (b) and CNN output (c)}
 \label{expRobotMockup}
\end{figure}

Finally, we performed the experiments on the real robot walking over a customized rough terrain. The example results are presented in Fig.~\ref{expRobotMockup}. The obtained elevation map (Fig.~\ref{expRobotMockup}b) is less accurate than the map obtained in the simulation experiments due to noise, but the robot can still identify risky edges and place its feet on the stable positions (see supplementary video).

\section{Conclusions and Future Work}
\noindent
In this paper, we propose a novel foothold selection method for legged systems. In contrast to methods known from the literature, the proposed method learns a model that evaluates the terrain patches and check all constraints in a single step. The time complexity for the inference is significantly reduced. With the proposed method, the robot avoids placing its feet on the edges or steep slopes. The neural network also implicitly takes into account the kinematic range of the leg and detects self-collisions and collisions with the ground. The proposed foothold selection module is integrated with the controller of the robot. In the simulation and experiments with the real robot, we present the properties and the efficiency of the proposed method.

In future, we plan to use the neural network to optimize the foothold position and the posture of the robot in a single step taking into account of the environment state (terrain model), kinematic, collision and stability constraints.

\end{document}